\documentclass[sigconf,nonacm]{acmart}
\AtBeginDocument{%
  \providecommand\BibTeX{{%
    \normalfont B\kern-0.5em{\scshape i\kern-0.25em b}\kern-0.8em\TeX}}}

\setcopyright{acmcopyright}
\copyrightyear{2023}
\acmYear{2023}
\begin{document}

\title{Considerations for Minimizing Data Collection Biases for Eliciting Natural Behavior in Human-Robot Interaction}

\author{Snehesh Shrestha}
\email{snehesh@umd.edu}
\orcid{1234-5678-9012}
\affiliation{%
  \institution{University of Maryland College Park}
  \city{College Park}
  \state{Marland}
  \country{USA}
  \postcode{20740}
}

\author{Ge Gao}
\email{gegao@umd.edu}
\orcid{0000-0003-0348-615X}
\affiliation{%
  \institution{University of Maryland College Park}
  \city{College Park}
  \state{Marland}
  \country{USA}
  \postcode{20740}
}

\author{Cornelia Fermuller}
\email{fermulcm@umd.edu}
\orcid{0000-0003-2044-2386}
\affiliation{%
  \institution{University of Maryland College Park}
  \city{College Park}
  \state{Marland}
  \country{USA}
  \postcode{20740}
}

\author{Yiannis Aloimonos}
\email{jyaloimo@umd.edu}
\orcid{0000-0002-8152-4281}
\affiliation{%
  \institution{University of Maryland College Park}
  \city{College Park}
  \state{Marland}
  \country{USA}
  \postcode{20740}
}

\renewcommand{\shortauthors}{Shrestha, et al.}

\begin{abstract}
Many of us researchers take extra measures to control for known-unknowns. However, unknown-unknowns can, at best, be negligible, but otherwise, they could produce unreliable data that might have dire consequences in real-life downstream applications. Human-Robot Interaction standards informed by empirical data could save us time and effort and provide us with the path toward the robots of the future. To this end, we share some of our pilot studies, lessons learned, and how they affected the outcome of our experiments. While these aspects might not be publishable in themselves, we hope our work might save time and effort for other researchers towards their research and serve as additional considerations for discussion at the workshop.
\end{abstract}

\keywords{hri, human factor, robots, data collection, standards, pilot studies}

\begin{teaserfigure}
  \includegraphics[width=\textwidth]{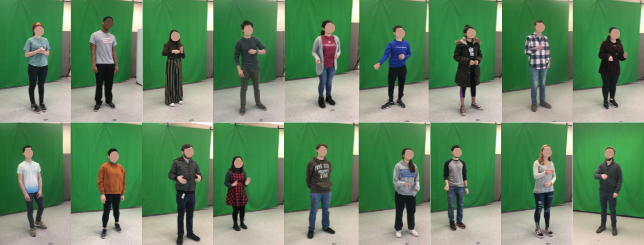}
  \caption{A sample of participants with diverse hand gestures while they command the robot to cut onions.}
  \label{fig:teaser}
\end{teaserfigure}
\maketitle

\section{Extended Abstract}
Natural human-robot interaction data can help robots learn from signals that are unstructured, mixed-modal, and consists of implied contexts. In our experiments, we find that they contain contradictory phrases and repair mechanisms. To incite natural human behavior in the lab is challenging. We took inspiration from prior literature, conducted several pilot studies, and developed a Wizard of Oz study (WoZ) experiment design to incite natural emergent human behavior.

To achieve the best of both worlds (in the wild and controlled lab study), our WoZ experiment design deceived the participants into believing the remote-controlled robot was fully autonomous. In this extended abstract, we will discuss a number of selected factors that we considered. These are some pilot studies we conducted that could affect participant behavior to validate independent and dependent control variables as well as the workflow. We will discuss the following considerations whose findings informed our experiment design decisions.
\begin{itemize}
    \item The effect of experiment \textit{instructions}
    \item The \textit{WoZ clues} that participants might be able to use to figure out the hidden agenda
    \item The \textit{priming} effects from \textit{practice} sessions
    \item The \textit{background noise}, and
    \item The robot's \textit{appearance} and \textit{identity}
\end{itemize}

\subsection{Instructions}
\label{appen:instructions_human}
We tested various modalities for our applications based on the recommendations \cite{Fothergill2012-sb}. Our findings in our pilot studies were in line with \cite{Fothergill2012-sb, Charbonneau2011-yi}, where the instruction modality significantly impacted the participants' behavior. For instances when text instructions were provided similar to \cite{Cauchard2015-vv}, participants preferred speech and used the exact words for the action and the object with little or no gestures. With videos of people performing the task similar to \cite{Charbonneau2011-yi}, participants copied the exact style of the demonstration of the actor. The one with the most variance in speech vocabulary and styles of gestures was when we showed before-after video clips to show the pre-task and post-task states; for example, to turn on a stove, we showed a zoomed-in video of a stove that was turned off and faded out to a video of the stove with the fire burning. For cutting an apple, a video of a whole apple on a cutting board being approached by a knife and faded into the apple that was cut into pieces where the knife is leaving the screen. And these videos were repeated in a loop with a 1-second gap. With such substantial differences in behavioral outcomes, we believe the instruction mechanisms used in the experiments and data collection sessions could benefit from guidelines and standardization.

\subsection{WoZ Clues}
\label{appen:woz_cues}
Wizard of Oz study (WoZ) is research experiment method where participants interact with a system the participants believe to be autonomous, however, the system is being operation fully or partially by another human\cite{dahlback1993wizard, riek2012wizard}. However, in WoZ studies, people can intuitively figure out the patterns, such as key press and mouse click sounds corresponding to robot actions. We experimented with masking the actual clicks and key presses with random ones. However, in the post-interview, the pilot test participants still seem to be able to figure out that researchers might be controlling the robot. So we created soft rubber remote control keys that use an IR receiver using an Arduino micro-controller USB adapter to send keys to the WoZ UI with virtually no sound that the researcher kept in their pocket. With this implementation, during the experiment, the researchers made sure when the experiment was being conducted, they did not sit at the control computer and appear to be moving around doing other things, appearing busy, staring at their phone, seemingly distracted, or looking at the participants showing attention in making sure the system was working without any technical issues. With this implementation, all of our participants believed that the robot was acting independently, and none suspected the WoZ setup to be a possibility.

\begin{figure}
  \includegraphics[width=\linewidth]{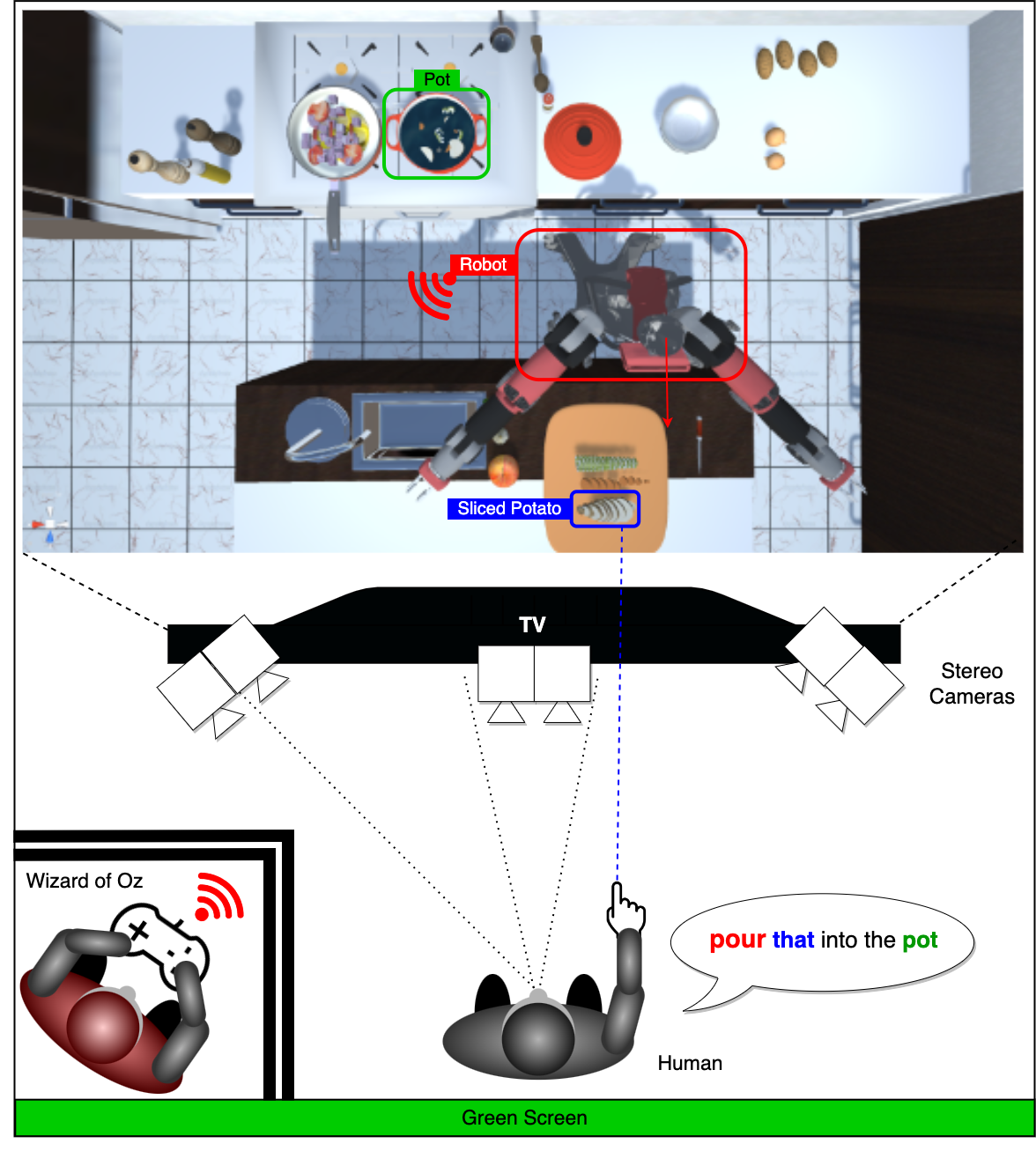}
  \caption{The researcher (WoZ) shown here in the bottom-left box is controlling the robot hidden from the participant, who believes the robot is autonomous.}
  \label{fig:setup}
\end{figure}

\subsection{Practice Session}
\label{appen:practice_session}
Practice sessions, especially performed right before the experiments, can strongly impact the outcome of the participant's behavior. For example, in our experiment, it was essential to ensure that participants were not primed to use one modality versus the other. So steps were taken to design the session with a mixture of related and unrelated commands where both speech and gestures were used to command the robots. If participants used a single modality only, they were encouraged to test out using the other modality. Participants interacted with the robot and asked researchers questions during practice. Once the practice was completed, participants were not allowed to interact with anyone other than the robot, even if they had questions or felt stuck as they were told that the experiment was designed for them to experience such scenarios and had to use creative methods to make the robot understand what they wanted the robot to do.

\subsection{Background Noise}
\label{appen:background_noise}
One hypothesis was that background noise can cause people to use more gestures. We considered three types of noise recording playback (lawn mower, people talking, and music) but only tested with people talking as background noise as that was the only example people found to be believable and not simulated. We tested three sets of loudness (M (dB) = 58, 63, 70, SD = 10, 13, 15). In our study (N=8), from people's use of speech and gesture and the post-interview, we found that (a) people tune out the background noise instead of using more gestures, (b) people wait for gaps of silence or lower-level noise in cases of speech or periodic noise, and (c) the noise had to be so loud that none of the speech could be heard for them to use gestures instead of speech. For these reasons, we decided not to use background noise as an independent variable. However, as a future possible directions, with a visualization of robot's perception of the sound, for example, robot picking up words from the background noise, or participant's speech being drowned out by the background noise, participants might use more gestures.

\subsection{Robot Appearance and Identify}
\label{appen:robot_face_and_name}
To reduce the effect of perceived gender, age, and personality by manipulating facial attributes, we considered the 17 face dimensions based on \cite{Kalegina2018-ki} study to design the face of the robot to be the most neutral face. The mouth of the robot was removed as not having a mouth did not have a significant adverse effect on the neutral perception of the robot. Having a mouth gave people the idea that the robot could speak, potentially causing the participant to prefer speech over gesture. To appear dynamic, friendly, and intelligent, we made the robot blink randomly between 12 and 18 blinks per minute \cite{Takashima2008-tz} with ease-in and ease-out motion profile \cite{Trutoiu2011-se, thomas1995illusion}. We further conducted pilot tests to analyze the head nod motions (velocity and the number of nods) and facial expressions for confusion expression. Additionally, we avoided using gender-specific pronouns ``he/him" and ``she/her" and referred to the robot as ``the robot" or ``Baxter," which is also the manufacturer-given name printed on the robot body that tends to be used both as a male and female name \cite{BaxterName_Wikipedia-qz}.

\bibliographystyle{ACM-Reference-Format}
\bibliography{HRI2023-Submission}

\end{document}